# A VLSI DESIGN AND IMPLEMENTATION FOR A REAL-TIME APPROXIMATE REASONING

Masaki Togai
Rockwell International Science Center
Thousand Oaks, CA 91360

and

Hiroyuki Watanabe
AT&T Bell Laboratories
Holmdel, New Jersey 07733

ABSTRACT. The role of inferencing with uncertainty is becoming more important in rule-based expert systems (ES), since knowledge given by a human expert is often uncertain or imprecise. We have succeeded in designing a VLSI chip which can perform an entire inference process based on fuzzy logic. The design of the VLSI fuzzy inference engine emphasizes simplicity, extensibility, and efficiency (operational speed and layout area). It is fabricated in 2.5 µm CMOS technology. The inference engine consists of three major components; a rule set memory, an inference processor, and a controller. In this implementation, a rule set memory is realized by a read only memory (ROM). The controller consists of two counters. In the inference processor, one data path is laid out for each rule. The number of the inference rule can be increased adding more data paths to the inference processor. All rules are executed in parallel, but each rule is processed serially. The logical structure of fuzzy inference proposed in the current paper maps nicely onto the VLSI structure.

A two-phase nonoverlapping clocking scheme is used. Timing tests indicate that the inference engine can operate at approximately 20.8 MHz. This translates to an execution speed of approximately 80,000 Fuzzy Logical Inferences Per Second (FLIPS), and indicates that the inference engine is suitable for a demanding real-time application. The potential applications include decision-making in the area of command and control for intelligent robot systems, process control, missile and aircraft guidance, and other high performance machines.

1. Introduction.

One of the first expert system (ES) developments was **fuzzy logic** or a **rule-based** control system developed in 1974 by Mamdani and Assilian [1]. The system accepted human knowledge of control strategies expressed verbally, or by **linguistic values**, and encoded it directly as computer programs. The controlled system was a small steam engine. The verbal rules were of the form shown as follows:

    IF    the pressure error is positive and big, and the change in pressure error is not negative medium or big
    THEN  make the heat change negative and big.

The distinguishing feature of fuzzy logic control is that it models the expert human operator instead of the process. In many practical cases, no model of the process exists because the process is too complex and ill-understood to be modelled mathematically. For such processes, a skilled human controller may be able to operate the plant successfully. The operators are often capable of expressing their operating practice in the form of rules, which may be programmed into a rule-based controller.

Fuzzy control systems and expert systems have one thing in common: both want to model human experience, human decision-making behavior. Fuzzy logic has been successfully incorporated in several fuzzy control systems and expert systems [3-5]. Fuzzy inference is also proposed in real-time decision-making in the area of command and control [6] to select the most suitable guidance and algorithm for intercepting missiles. Selection is done by considering a constantly changing environment; that is, the relative angular positions, accelerations, and distances of an evader and a missile. These examples show the need for an efficient inference engine to handle large rule sets and for real-time use.

We have designed, fabricated and tested a VLSI chip which can perform fuzzy inference (or **approximate reasoning**) in real-time. The chip is designed to be extensible to handle a large number of rules without slowing process speed.

## 2. Execution of Approximate Reasoning.

### 2.1 Fuzzy Variables.

In a fuzzy control system, a fuzzy implication concerned with a control input to a process is called a **fuzzy control rule**, as shown in the previous section. This rule is more concisely written as

    IF E=PB and CE=7(NM or NB) THEN CO=NB   ,

where E is the error, CE is the change of error, and CO is the controller output, i.e., the control input to the process. The variables PB = positive and big, NM = negative and medium and so on are called **fuzzy variables**. Fuzzy variables are usually chosen as shown in Figure 2.

In Figure 1, a fuzzy variable A is represented by a fuzzy subset, i.e., a set of ordered pairs of an element, $u_i$, of A and its **grade of membership**:

$$A = \{(u_i, \mu_A(u_i))\} \quad , \quad u_i \epsilon U \quad , \tag{1}$$

where U is a universe of discourse.

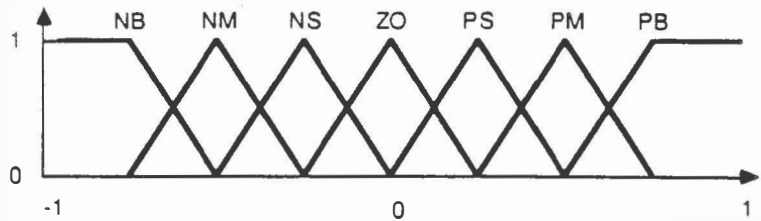

Figure 1.  Typical fuzzy variables.

2.2 <u>Approximate Reasoning</u>.

In the traditional logic the main tools of reasoning are tautologies such as the **modus ponens**, i.e.:

| | |
|---|---|
| Premise | A is true |
| Implication | If A then C |
| Conclusion | C is true |

A and C are statements or propositions (crisply defined) and the C in the conditional statement is identical to the C of the conclusion.  Two quite obvious generalizations of the **modus ponens** are:

1. To allow statements which are characterized by fuzzy subsets.

2. To relax (slightly) the identity of the "C's" in the implication and the conclusion.

This version of the modus ponens is called "generalized modus ponens," or an "approximate extension" of **modus ponens** [7].

Let A, A', C, C' be fuzzy statements this time.  Then the **generalized modus ponens** reads:

| | |
|---|---|
| Premise | x is A' |
| Implication | If x is A then z is C |
| Conclusion | z is C' |

For instance,

| | |
|---|---|
| Premise | This tomato is very red |
| Implication | If a tomato is red then the tomato is ripe |
| Conclusion | This tomato is very ripe |

In 1973, Zadeh suggested the **compositional rule of inference** for approximate reasoning.

291

Suppose we have the two implications or rules with two fuzzy variables in IF part and one in THEN part:

Rule 1: If x is $A_1$, y is $B_1$ then z is $C_2$ ,

Rule 2: If x is $A_2$, y is $B_2$ then z is $C_2$ .

Then given observation A' and B', the weights, i.e., w1 and w2, of the premises are calculated by:

$$w_1 = \min(\alpha_1^A, \alpha_1^B) \quad , \tag{2}$$

$$w_2 = \min(\alpha_2^A, \alpha_2^B) \quad , \tag{3}$$

where

$$\alpha_i^A = \max_x (A' \cap A_i) \tag{4}$$

$$\alpha_i^B = \max_y (B' \cap B_i) \tag{5}$$

The first rule is $w_1 \cap C_1$; the second is $w_2 \cap C_2$. Then overall resulting conclusion C' is obtained by

$$C' = (w_1 \cap C_1) \cup (w_2 \cap C_2) \quad .$$

The reasoning process is illustrated in Figure 2.

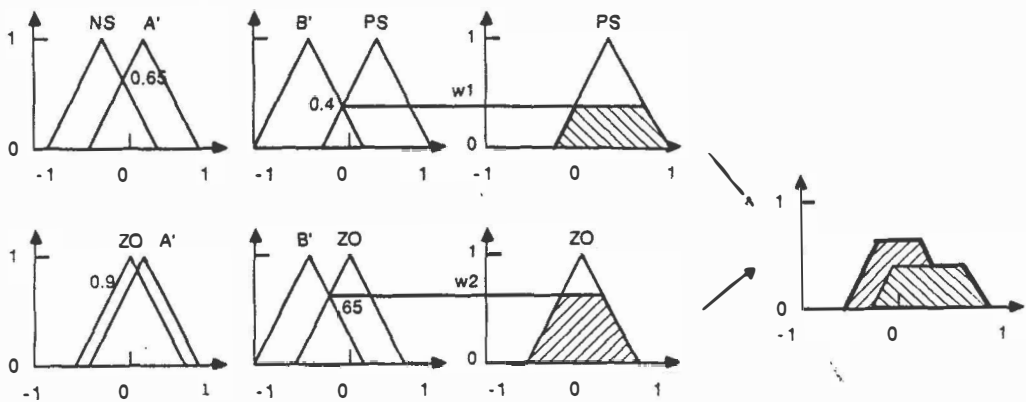

Figure 2. Approximate reasoning.

## 3. VLSI Design for Approximate Reasoning.

The VLSI inference engine consists of three major parts: a rule set memory, an inference processing unit, and a controller. The inference mechanism described in the previous section executes all the rules in parallel. Because of the very high rate of communication between the rule set storage unit and the inference processing unit, we decided to store the rules on a chip. Otherwise, the restriction of the pin count prevents the effective use of the parallelism.

### 3.1 Storage and Format of Rule Set.

The rule set can be stored using either a random access memory (RAM) or a read only memory (ROM). The advantage of using RAM is its flexibility. Depending on the application, the rule set can be loaded from off-chip. On the other hand, ROM takes much less area for the same amount of data and operates faster. The control unit of the inference engine can be very simple, since we do not need to load a rule set from off-chip.

We considered the size of the fuzzy variables subset and the grade of fuzziness for a practical use. In most cases, a fuzzy variable has three to sixteen elements; the grade of fuzziness, three to twelve (e.g., [4,8]). For this implementation, we can limit the universe of discourse of a fuzzy subset to be a finite set with 32 or 64 elements (i.e., 5 or 6 bits). The membership function is discretized in 16 levels (i.e., 4 bits). That is, 0 represents no membership, 15 represents a full membership and other numbers represent points in the unit interval [0,1], Therefore, a fuzzy subset can be digitized with 128 or 256 bits. The format of the rule representation is as follows:

Rule 1:  $A_i \rightarrow C_i$

where

| | $x_1$ | $x_2$ | $x_3$ | ... | $x_j$ | ... | $x_{32}$ |
|---|---|---|---|---|---|---|---|
| $A_i$: | 0010 | 0100 | 1111 | ... | $\mu_{A_i}(x_j)$ | ... | 0000 |

| | $z_1$ | $z_2$ | $z_3$ | ... | $z_j$ | ... | $z_{32}$ |
|---|---|---|---|---|---|---|---|
| $C_i$: | 0000 | 0001 | 0011 | ... | $\mu_{B_i}(z_j)$ | ... | 1100 |

Here, each 4 bits represents degree of membership for each element of the universe. For example, $u_1$ has a degree of membership 2/15 and $u_3$ has a full degree of membership in a subset $A_i$. Each 4 bits integer is stored most significant bit first. All the rules are accessed in parallel. An individual rule is, however, accessed in the serial manner. Two memory modules are used for storing the antecedent A's and B's, and the conclusion C's of the rule set, respectively.

293

## 3.2 Inference Processing Unit.

The inference processing unit consists of two basic logical circuits for minimum and maximum operations. These circuits are used to implement fuzzy intersection and fuzzy union operations. The circuit for minimum operation takes two integers and produces the smaller number; the circuit for maximum operation produces the larger number. These logical circuits process integers serially.

The basic data path of the inference engine for processing a single rule is shown in Figure 3. This directly corresponds to a single data path of the inference engine described in Figure 2. The shift register is used for keeping the maximum after the fuzzy intersection operation in the first level. This is necessary since within an individual rule operations are performed serially. The register records the value of the maximum point, a value $\alpha_i$, when the first level has finished its operation. The last operation of the second level requires taking the maximum membership function over all the data paths (i.e., all the rules). This operation is accomplished by connecting the maximum units in the binary tree structure.

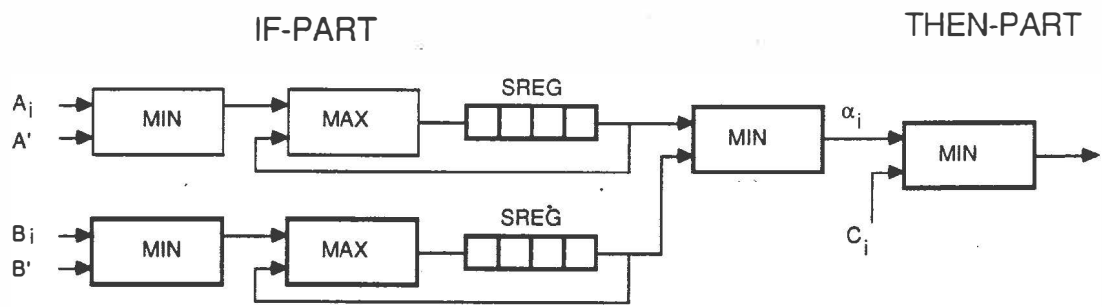

Figure 3. Single data path of inference processor.

## 3.3 Controller.

Because of the simplicity of the architecture, the controller of the inference engine is straightforward. It consists of two counters for accessing two memory modules serially. The controller generates a reset signal for the minimum and maximum elements for every 4 cycles. The controller starts to access the conclusion parts of the rules as soon as the processing of the antecedent parts are finished. It also notifies a user the beginning of the valid output.

## 4. AT&T Chip.

The first implementation of this technology was demonstrated at AT&T Bell Laboratories [9]. It stores 16 rules of the form: if A then C. Fuzzy variables, i.e., A's and C's, have 31 elements with 16 levels of membership. The premise A' and the conclusion C' are also presented by the same format; they are loaded and produced serially.

A nonoverlapping two-phase clocking scheme, supplied from off-chip, is used. The operation is initiated by a reset signal that must last one clock cycle and resets the entire circuit. Input of an observation should be started two clock cycles after the reset signal; that is, on the third clock cycle. The inference engine begins to produce the result on the 133rd cycle after the reset signal. The beginning of the valid output is signaled by the controller. The active size area is 2.99 × 3.48 mm. A 68-pin package is used. A microphotograph of the fabricated chip is shown in Figure 4.

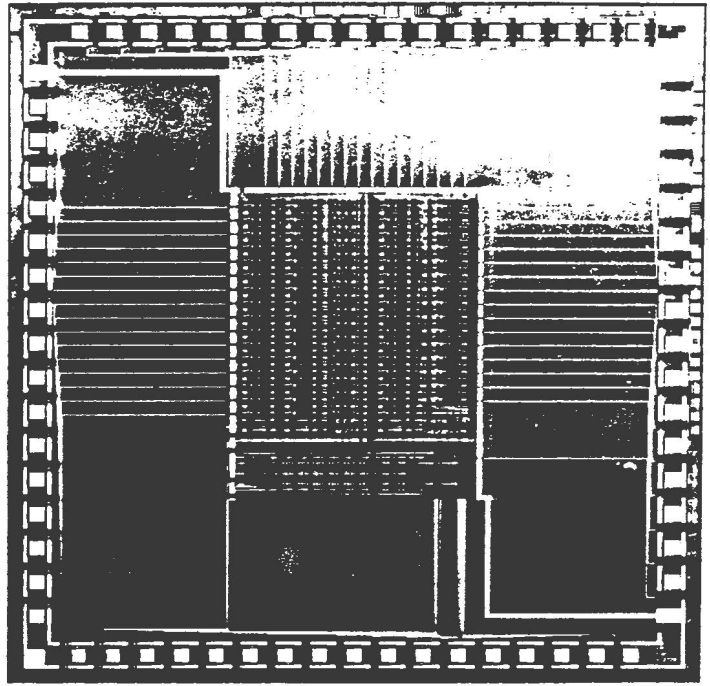

Figure 4. Microphotograph of fuzzy inference chip at AT&T Bell Laboratories.

Timing tests on fabricated chips indicated a 20.8 MHz (48 ns cycle time) operating rate. With the current data format of 124 bits per rule, a single inference process takes 256 clock cycles. Thus, the engine can perform approximately 80,000 Fuzzy Logical Inference Per Second (FLIPS), which is more than 1,000 times faster than the simulation conducted on the VAX-11/785. The chip has a drastic speed advantage over the conventional approach.

5. Extensibility of the Design.

The advantage of the architecture is its simplicity and extensibility. The format of the fuzzy variables representation can be expanded with minor changes in the layout. For example, the number of elements to represent a fuzzy variable can be either increased or decreased by minor modification of the controller. The resolution of membership function can be modified by both changing the length of the shift register and modifying the controller. The number of rules can be increased by laying out more data paths and modifying the binary tree layout accordingly.

6. Conclusion.

VLSI technology opens a new dimension of AI research. A chip for approximate reasoning is designed using CMOS technology and fabricated. The total response time of the chip is independent to the size of rules. It can be used in many real-time applications such as command and control, machine vision, system diagnosis, and factory automation.

It is feasible in the near future to build a chip which can perform over 1 Mega fuzzy inference per second (FLIPS). This is the first step towards a fuzzy computer, which could be a **Mega FLIPS Machine.**

7. References.

[1] E.H. Mamdani and S. Assilian, "A Case Study on the Application of Fuzzy Set Theory to Automatic Control," Proc. IFAC Stochastic Control Symp. Budapest, 1974.

[2] E.H. Mamdani, J.J. Ostergaard and E. Lembessis, "Use of Fuzzy Logic for Implementing the Rule-Based Control of Industrial Processes," in Fuzzy Sets and Decision Analysis (H.J. Zimmermann, L.A. Zadek and B.R. Gains, eds.) TIMS Series Studies in the Management Sciences, 20, North-Holland, Amsterdam, 1984.

[3] P.P. Bonissone and H.E. Johnson, Jr., "Expert System for Diesel/ Electric Locomotive Repair," Human Systems Management 4, pp 225-262, 1984.

[4] M. Sugeno and K. Murakami, "Fuzzy Parking Control of Model Car," Proc. the 23rd IEEE Conf. Decision and Control (December 12-14, Las Vegas, NE) 1984.

[5] L.A. Zadeh, "The Rule of Fuzzy Logic in the Management of Uncertainty in Expert Systems," Fuzzy Sets and Systems 11, pp. 119-227, 1983.

[6] K. Kawano, M. Kosaka and S. Miyamoto, "An Algorithm Selection Method Using Fuzzy Decision-Making Approach," Trans. Society of Instrument and Control Engineers 20, No. 12, pp. 42-49, 1984 (in Japanese).

[7] L.A. Zadeh, "Outline of a New Approach to the Analysis of Complex Systems and Decision Processes," IEEE Trans. Systems Man, and Cybernetics, SMC-3, pp. 28-45, 1973.

[8] L.I. Larkin, "A Fuzzy Logic Controller for Aircraft Flight Control," Proc. the 23rd IEEE Conf. Decision and Control (December 12-14, Las Vegas, NE), 1984.

[9] M. Togai and H. Watanabee, "A VLSI Implementation of a Fuzzy Inference Engine: Toward an Expert System on a Chip," Information Sciences, 38, 147-163 (1986).